\documentclass[journal]{IEEEtran}
\usepackage{graphicx}
\usepackage{epstopdf}
\usepackage[cmex10]{amsmath}
\usepackage{xcolor}
\usepackage{eqnarray}
\usepackage{latexsym}
\usepackage[caption=false,font=footnotesize]{subfig}
\usepackage{fixltx2e}
\usepackage{color}
\usepackage{amsfonts}
\usepackage{cite}

\begin{document}
	\bstctlcite{IEEEexample:BSTcontrol}
\newtheorem{michaeldo1}{Proposition} 
\newtheorem{michaeldo2}{Theorem} 
\title{Performance Control of Tendon-Driven Endoscopic Surgical Robots With Friction and Hysteresis}
\author{Thanh Nho~Do,
        Tegoeh~Tjahjowidodo,
        Michael Wai Shing~Lau,
        and~Soo Jay~Phee
}


\markboth{}
{Shell \MakeLowercase{\textit{et al.}}: Bare Demo of IEEEtran.cls for Journals}

\maketitle
\begin{abstract}
In this study, a new position control scheme for the tendon-sheath mechanism (TSM) which is used in flexible medical devices is presented. TSM is widely used in dexterous robotic applications because it can flexibly work in limited space, in constrained environments, and provides efficient power transmission from the external actuator to the distal joint. However, nonlinearities from friction and backlash hysteresis between the tendon and the sheath pose challenges in achieving precise position controls of the end effector. Previous studies on the TSM only address the control problem under the assumptions of known tendon-sheath configuration and known model parameters of the backlash hysteresis nonlinearity. These approaches can have adverse impact and limitations on the overall system performances and practical implementation. This paper presents a new approach to model and control the TSM-driven flexible robotic systems. The designed controller does not require exact knowledge of nonlinear friction and backlash hysteresis parameters, only their bounds are online estimated. Simulation and experimental validation results show that the proposed control scheme can significantly improve the tracking performances without the presence of the exact knowlege of the model parameters and the sheath configuration.

\end{abstract}

\section{Introduction}

\IEEEPARstart{A}{dvances} in Minimally Invasive Surgery (MIS) allow surgeons to perform efficient, high fidelity, and more complex tasks for surgical treatment while avoiding unexpected damages such as tissue trauma and reducing recovery time after the surgery. One of the great developments of MIS is the Natural Orifice Transluminal Endocopic Surgery (NOTES). In NOTES, procedures are performed through transvaginal, transgastric, or transesophageal approaches with potentially scar-free surgery, better cosmetics, and less pain \cite{Vitiello6392862}, \cite{Le2016323}. In order to achieve the true potential of NOTES, highly dexterous and cost effective instruments are desired. These desires can be satisfied with the use of tendon-sheath mechanism (TSM). The tendon can pass through a flexible sheath and it is able to work in very narrow paths and constrained environments with high payload and efficient transmission. In most of flexible endoscopic systems, robotic catheter, and robotic hands, a pair of TSMs is often used to actuate the distal joint while the actuator is externally provided \cite{6269103}, \cite{6663692}, \cite{6661461}, \cite{TepRA7219674},. However, because the nature of the mechanical design for the TSM creates nonlinear friction and backlash hysteresis, it is a challenging task to acurately model and control the position of the distal joint, especially in the absence of model parameters and the sheath configuration. These nonlinearities have also restricted the use of the TSM in the development of various flexible endoscopic devices despite its advantages \cite{6678696}, \cite{Yeung2012345}, \cite{7299273}, \cite{7274578}. 

Friction is a nature resistance to relative motion between two contacting bodies. If the position controller is designed without considering the friction and backlash hysteresis, the tracking error in the closed-loop system will result in steady-state or oscillations \cite{5411804}, \cite{TriVo5382520}, \cite{Tjahjowidodo2007959}. Many studies have been devoted to the development of position/force transmission and control for the TSM. Based on the system dynamics, the tendon-sheath transmission can be either modelled by an approximation of the backlash hysteresis profile \cite{AgrawalHysteresisCom}, \cite{Do201412} or in terms of the lumped-mass model in combination with static friction model like the Coulomb model \cite{agrawalIEEETrans}. It has been known that the use of static friction model can result in discontinuous problem when the system reverses its motion. Hence, Palli \textit{et al.} \cite{palli2012modelingidentification} used the LuGre model in combination with the lumped-mass parameters to overcome this drawback. However, the model approach is still not able to capture the complete nonlinear characteristics of tendon-sheath friction for both small displacement (presliding regime) and  large displacement (sliding regime). In addition, the tendon configuration and force feedback must be known in advance. To fully capture the dynamic properties for both the sliding and presliding regimes, Do \textit{et al.} \cite{DoInvestigation}, \cite{doICINCO2013}, \cite{DoRCIM2015} first designed novel dynamic friction models for a single TSM. Although the designed models were able to characterize all the friction phenomena for the TSM, these models require a large number of parameters in their structure. Later, Do \textit{et al.} \cite{DoDUbaijournal} developed a new dynamic friction model for a pair of TSMs with fewer parameters in the designed approach. However, no motion control schemes have been introduced to compensate the tracking error.

Many attempts on the position control of the TSM have been studied in the literature. Agrawal \textit{et al.} \cite{AgrawalHysteresisCom} used a smooth backlash inverse model to enhance the tracking performances for the tendon-sheath system. In the applications of robotic catheter and flexible endoscope, Bardou \textit{et al.} \cite{BardouImprovement}, Kesner \textit{et al.} \cite{KesnerIEEE}, and Reilink \textit{et al.} \cite{reilink2013image} used an inverse backlash model to compensate for the phase lags between the output position and the desired trajectory of the TSM. However, an inverse model and a discontinuous switching function of velocity were needed. Although Do \textit{et al.} \cite{Do201412}, \cite{doCISRAM2015}, \cite{doICINCO2014} used a direct inverse model-based feedforward compensation to overcome the above limitations, the controller was designed for a fixed tendon-sheath configuration, known backlash hysteresis parameters, and known bounds in advance. 

Several work has been introduced to compensate for the tracking error in the presence of nonlinear friction and backlash hysteresis model by using different nonlinear control techniques. Recent studies on the mechanical systems like magnetic devices or piezoelectric actuators have focused on the perfect cancellation of nonlinearities using inverse models and adaptive algorithms with unknown model parameters and disturbances. However, little experimental validations have been carried out to demonstrate the practical implementation for the proposed approaches \cite{hassani2014survey}, \cite{6701387}, \cite{6022799}. To fulfil the complete picture for the tendon-sheath control, this paper introduces an adaptive compensation technique for the TSM using the new dynamic friction model given by \cite{DoDUbaijournal}. The designed approach takes into account the dynamics of the tendon routing through the sheath regardles of the sheath configuration. In addition, the designed laws do not require the knowledge of the friction model and environment parameters; the structure of the controller is designed based on the standard backstepping technique and adaptive updated laws \cite{hassankhalil}, \cite{KristicKokotovic}\cite{Do201567}, \cite{DoMSSP2015}, \cite{DoMechanism}, \cite{doTLuan2014}. To validate the proposed model approach, an experimental platform is introduced. Simulation and real-time test results confirm that the proposed controller significantly improves the position tracking performances for the TSM.

The rests of this paper are organized as follow: In section II, problem formulation for the tendon-sheath system is given. Section III introduces the nonlinear and adaptive controller design. Simulation illustration for the proposed control scheme is shown in section IV. The experimental setup and validation results are shown in section V. Finally, the discussion and conclusion are drawn by section VI.

\section{Problem formulation}
\subsection{Dynamic Friction for the TSM}
The dynamic friction force $F$ for the TSM can be described as follows \cite{DoDUbaijournal}:
\begin{equation}
\setlength{\nulldelimiterspace}{0 pt}
\left\{
\begin{IEEEeqnarraybox}[\relax][c]{l's}
F=k_x\phi x_i+k_\zeta \zeta+\upsilon\dot{x}_i+F_0 \\
\dot{\zeta}=\rho(\dot{x}_i-\sigma|\dot{x}_i||\zeta|^{n-1}\zeta+(\sigma -1)\dot{x}_i|\zeta|^n)\\
\phi=\frac{(e^{2\dot{x}_i}+\tanh(x_i)\tanh(\ddot{x}_i))}{e^{2\dot{x}_i}+1}
\end{IEEEeqnarraybox} \right.
\label{eq1}
\end{equation}

where $x_i, \dot{x}_i, \ddot{x}_i$ are the angular position (in radians), velocity, and acceleration at the actuator side, respectively; $F$ is the total friction forces in the TSM; $ k_\zeta$ is a factor that express the ratio of the internal state $\zeta$ to the friction force; $k_x$ is the stiffness factor that controls seperate curves of the hysteresis loops; $\rho>0, \sigma>0.5, n\geq 1$ are coefficients that control the shape and size of hysteresis loops for the friction force; $\tanh(\bullet)$ is the hyperbolic tangent function which is defined as $\tanh(\bullet)=(e^{2(\bullet)}-1)/(e^{2(\bullet)}+1)$; $\upsilon$ is the viscous coefficient; and $F_0$ is a offset point of the friction force; the dot at the top of variables represents the first derivative with respect to time.

\textit{Remark} 1: It is noted that the variable $\zeta$ given by (\ref{eq1}) is uniformly bounded for any piecewise continuous signals $x_i, \dot{x}_i$ (See \cite{ikhouane2007dynamic}, \cite{ikhouane2005adaptive} for more details). This important property will be used for the controller design in the next sections.

\subsection{Dynamic Model of the Distal Joint}
\begin{figure}[h]
\centering
\includegraphics[width=0.4\textwidth]{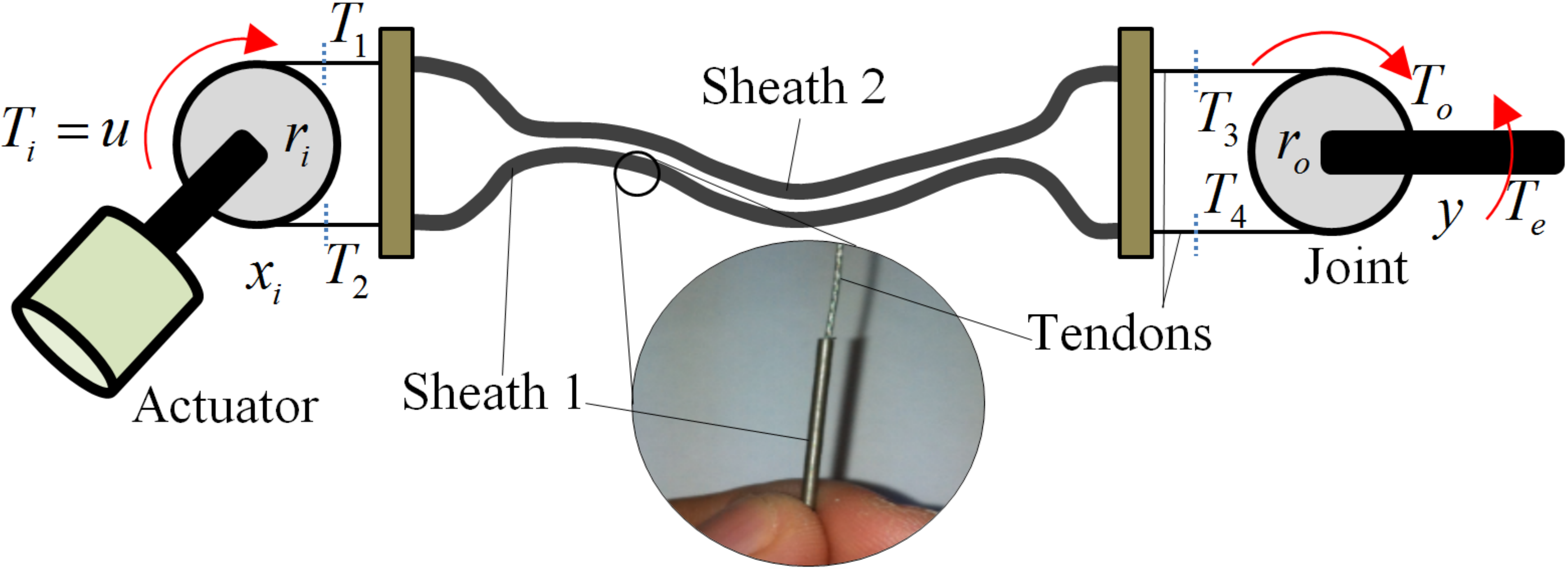}
\caption{Torque Transmission for the TSM}
\label{figure1}
\end{figure}

The dynamic model for the TSM-driven joint with the rotational inertias $m$, damping coefficient $c$, input torque at the distal end $T_o$, external disturbance torque $T_d$ with unknown bound, and environmental torque $T_e$ applied to the joint as shown in Fig. \ref{figure1} can be described as \cite{Mahvash4399954}:
\begin{align}
m\ddot{y}+c\dot{y}=T_o-T_e-T_d \label{eq2}
\end{align}

where $y$ is the rotational displacement of the distal joint.

Let $T_j (j=1,...,4)$ be tensions at the actuator and the joint sides of each tendons and let $r_i, r_o$ be radii of the pulleys at the actuator and joint sides. The mass of tendons is usually small and can be ignored. Then the relation between the proximal torque $T_i=u=(T_2-T_1)r_i$ ($u$ is the control input for the system) and the distal torque $T_o=(T_4-T_3)r_o$  can be expressed as $T_i/r_i=T_o/r_o+F$ where $F=(T_2-T_4)+(T_3-T_1)$ is the total friction forces between the tendons and the sheaths. The Eq. (\ref{eq2}) can be rewritten by:
\begin{align}
m\ddot{y}+c\dot{y}=r_o(u/r_i-F)-T_e-T_d \label{eq3}
\end{align}

We define the system states as $x=[x_1, x_2]^T=[y,\dot{y}]^T$; then the dynamics of the joint in the tendon-sheath system can be simplified as:
\begin{equation}
\setlength{\nulldelimiterspace}{0 pt}
\left\{
\begin{IEEEeqnarraybox}[\relax][c]{l's}
\dot{x}_1=x_2\\
\dot{x}_2=(1/m)(r_o(u/r_i-F)-T_e-T_d-cx_2)\\
y=x_1
\end{IEEEeqnarraybox} \right.
\label{eq4}
\end{equation}

The friction force $F$ is described by (\ref{eq1}). Then the variable $\dot{x}_2$ given by \ref{eq4} can be rewriten as:
\begin{align}
\dot{x}_2&=\frac{1}{m}(r_o(u/r_i-k_x\phi x_i-\upsilon\dot{x}_i)-cx_2+D)\nonumber\\
&=\frac{1}{m}((r_o/r_i)u+D)+\Theta^T\varphi
\label{eq5}
\end{align}

where $\varphi,D,\Theta$ are expressed with the following forms: $\varphi=[\phi x_i, \dot{x}_i, x_2]^T$=$[x_i (e^{2\dot{x}_i}+\tanh(x_i)\tanh(\ddot{x}_i))/(e^{2\dot{x}_i}+1), \dot{x}_i, \dot{y}]^T$, $D=-r_o (k_{\zeta}\zeta+F_0)-T_e-T_d$, and $\Theta=[-r_ok_x/m,-r_o\upsilon/m, -c/m]^T$.

The objective of this paper is to design the control input $u$ such that:
\begin{itemize}
\item The output $y$ accurately tracks the desired trajectory $y_r$ as close as possible.
\item The tracking error $e_r=y-y_r$ and estimate variables are bounded.
\end{itemize}
\section{Controller design}
To design the controller $u$ and suitable adaptive laws, some assumptions are needed: $(i)$ The positions $y, x_i$ are measurable; $(ii)$ the desired trajectory $y_r, \dot{y}_r, \ddot{y}_r$ are continuous and bounded; $(iii)$ The environmental torque $T_e$ and external disturbance $T_d$ are bounded by unknown bounds. From \textit{Remark} 1 and assumption $(iii)$, the variable $D$ is bounded by an unknown variable $D^*$ $(|D| \leq D^*)$ due to $\zeta, T_e, T_d$ are bounded.

To use the backstepping technique \cite{hassankhalil}, \cite{KristicKokotovic}, the change of coordinate transformations is made to the system (\ref{eq4}) and (\ref{eq5}):
\begin{align}
\xi_1&=x_1-y_r\label{eq6}\\
\xi_2&=x_2-\dot{y}_r+\alpha_{v1}\xi_1 \label{eq7}
\end{align}

where $\alpha_{v1}$ is a positive designed parameter that will be determined later.

From (\ref{eq6}) and (\ref{eq7}), the new coordinates for the system (\ref{eq4}) and (\ref{eq5}) can be expressed by:
\begin{align}
\dot{\xi}_1&=\xi_2-\alpha_{v1}\xi_1\label{eq8}\\
\dot{\xi}_2&=\frac{1}{m}((r_o/r_i)u+D)+\Theta^T\varphi-\ddot{y}_r+\alpha_{v1}\dot{\xi}_1\label{eq9}
\end{align}

Let $\hat{D}^*, \hat{m}, \hat{\Theta}$ and $\tilde{D}^*=D^*-\hat{D}^*, \tilde{m}=m-\hat{m}, \tilde{\Theta}=\Theta-\hat{\Theta}$ be the estimates and error estimates of $D^*, m, \Theta$, respectively. Define a virtual control input $\bar{u}$ such that $u=\hat{m}\bar{u}-\hat{D}^*\tanh(\xi_2/\epsilon)$ where $\epsilon > 0$ will be determined later. From (\ref{eq8}) and (\ref{eq9}), parameter update laws $\hat{D}^*, \hat{m}, \hat{\Theta}$ and control input $u$ are designed as follow:
\begin{align}
u&=(r_i/r_o)(\hat{m}\bar{u}-\hat{D}^*\tanh(\xi_2/\epsilon))\label{eq10}\\
\bar{u}&=\ddot{y}_r-\alpha_{v1}\dot{\xi}_1-\xi_1-\alpha_{v2}\xi_2-\hat{\Theta}^T\varphi \label{eq11}\\
\dot{\hat{\Theta}}&=k_{\Theta}\xi_2\varphi-\sigma_1\hat{\Theta} \label{eq12}\\
\dot{\hat{m}}&=-k_m\xi_2\bar{u}-\sigma_2\hat{m} \label{eq13}\\
\dot{\hat{D}}^*&=k_D\xi_2\tanh(\xi_2/\epsilon)-\sigma_3\hat{D}^* \label{eq14}
\end{align}
where $\alpha_{v1}, \alpha_{v2}, k_{\Theta}, k_m, k_D, \sigma_i (i=1,2,3)$ are positive parameters that adjust the designed controller $u$ to force the tracking error $e_r=y-y_r$ approaches desired values.

With (\ref{eq10}) to (\ref{eq14}), the following theorem holds:

\textit{Theorem} 1: Consider the nonlinear system (\ref{eq8}) and (\ref{eq9}) with the designed controller $u$ from (\ref{eq10}) to (\ref{eq11}) and adaptive laws (\ref{eq12}) to (\ref{eq14}), the following statements hold:
\begin{enumerate}
\item The tracking error $e_r$ and adaptive laws $\hat{D}^*, \hat{m}, \hat{\Theta}$ are globally uniformly bounded.
\item In the presence of unknown model parameters and their bounds, the position tracking error converges to a desired compact region $\Omega=\{|e_r|\big{|}|e_r| \leq 2\sqrt{\varrho/\Psi}\}$.
\end{enumerate}
where $\Psi=(\sigma_1/2k_{\Theta})\Theta^T\Theta+0.2785\epsilon D^*/m+(\sigma_2/2k_m)m+(\sigma_3/ 2mk_D) (D^*)^2$; $\varrho=\text{min}\{2\alpha_{v1}, 2\alpha_{v2}, \sigma_1, \sigma_2, \sigma_3\}$. 

\begin{IEEEproof}
See appendix A.
\end{IEEEproof}

\textit{Remark} 2: For practical implementation of the proposed controller and adaptive laws given by (\ref{eq10}) to (\ref{eq14}), simulations have been carried out. Some guidelines are recommended to determine the designed control parameters:
\begin{itemize}
\item The parameters $\alpha_{v1}$ and $\alpha_{v2}$ determine the level of the convergence speed of $e_r$. Large values of $\alpha_{v1}$ and $\alpha_{v2}$ will lead to faster convergence of the tracking error. However, if these values are too large, it may cause chattering in the controller. High values of parameters $k_\Theta, k_m, k_D$ will improve the parameter adaptation speed and tracking performances. In addition, $\epsilon$ is used to guarantee the smoothness and avoid the discontinuity of the controller $u$ given by (\ref{eq10}). A small value of $\epsilon$ is desired. Therefore, we suggest to fix $\alpha_{v1}$ and $\alpha_{v2}$ at acceptable large values and $\epsilon$ at a small value and then increase the other parameters such as $k_\Theta, k_m, k_D$ until the desired tracking performances are achieved.
\item  The parameters $\sigma_1, \sigma_2, \sigma_3$ are used to prevent the estimate values of $\hat{\Theta}, \hat{m}, \hat{D}^*$ to be very large. However, large values of $\sigma_1, \sigma_2, \sigma_3$ may suppress the prevention. Hence, small values for $\sigma_1, \sigma_2, \sigma_3$ are expected.
\end{itemize}

\section{Simulation Results}
In this section, simulation studies which illustrate the performances of the proposed adaptive controller are presented. Various simulations will be carried out in order to find a good set of controller parameters. The dynamic friction model parameters are chosen based on experimental validation and offline identification from Do \textit{et al.} \cite{DoDUbaijournal} where $k_x=0.01083, \rho=54.658, n=2.0458, \sigma=1.58, k_{\zeta}=0.14368, \upsilon=0.02686, F_0=0.0099$. The parameters for the dynamic joint are selected as $m=0.0349, c=0.0105$. The environment is simulated as an eleastic spring $T_e=k_e y$ where $k_e=0.4185$, $y$ is the output position. The disturbance is chosen like $T_d=0.2\sin(0.2\pi t)$. The designed controller and updated law parameters are selected basing on the guidelines given by the \textit{Remark} 2. With this idea, we choose the control parameters as $\alpha_{v1}=10,~\alpha_{v2}=15,~k_{\Theta}=0.5,~k_m=0.5,~k_D=1,~\sigma_i=0.01~(i=1,2,3),\epsilon=0.05$. The pulley radii are 25mm. The initial condition for updated estimate parameters are chosen such as $\hat{\Theta}(0)=[0,0,0]^T,~\hat{m}(0)=0,~\hat{D}^*(0)=0$. The desired trajectory is selected as $y_r=0.4\sin(0.4\pi t)$.

\begin{figure}[h]
\centering
\includegraphics[width=0.36\textwidth]{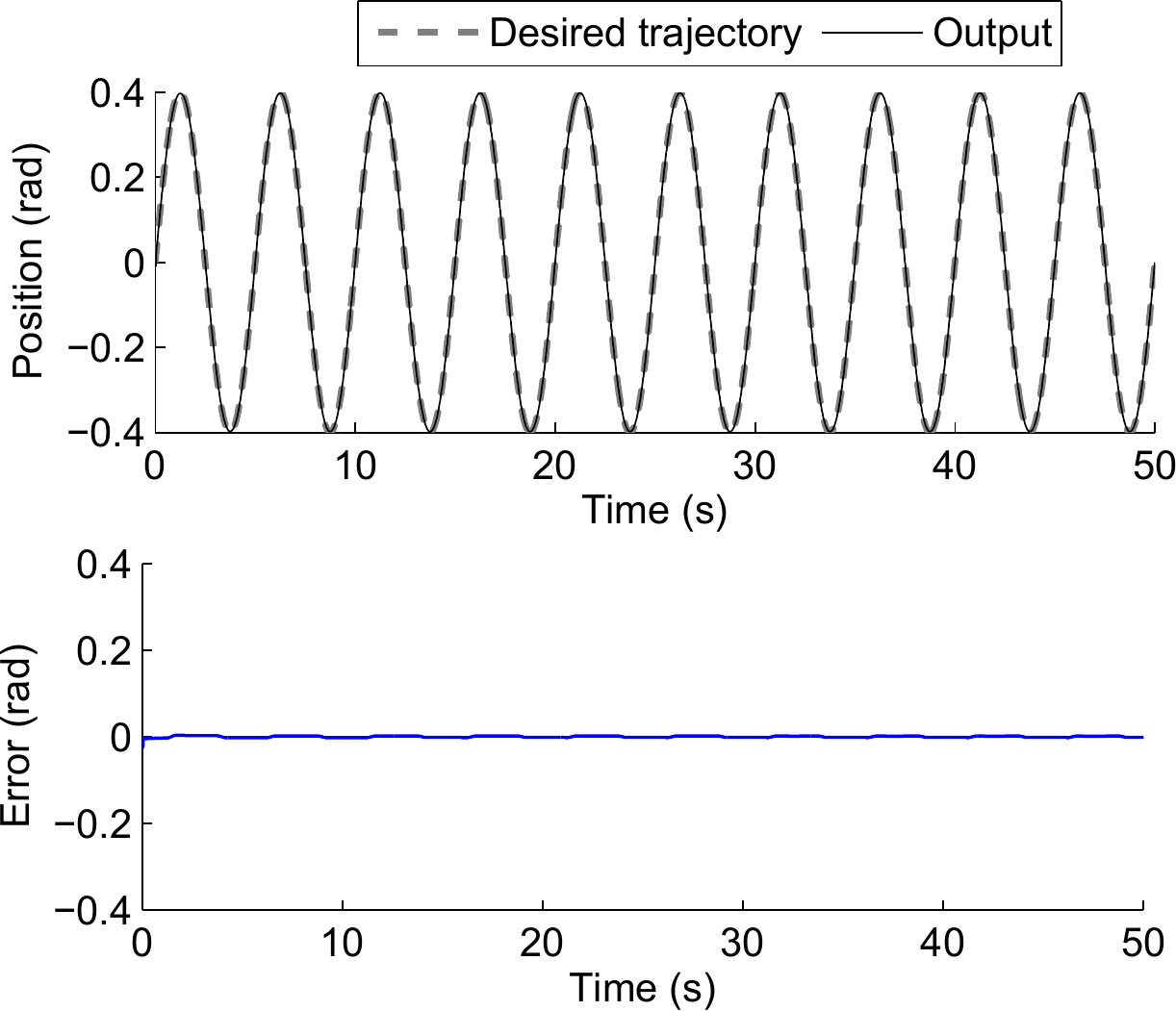}
\caption{Simulation results with the designed adaptive laws:(Upper panel) Time history of the desired trajectory $y_r$ and the output $y$; (Lower panel) Tracking error}
\label{figure2}
\end{figure}

\begin{figure}[h]
\centering
\includegraphics[width=0.36\textwidth]{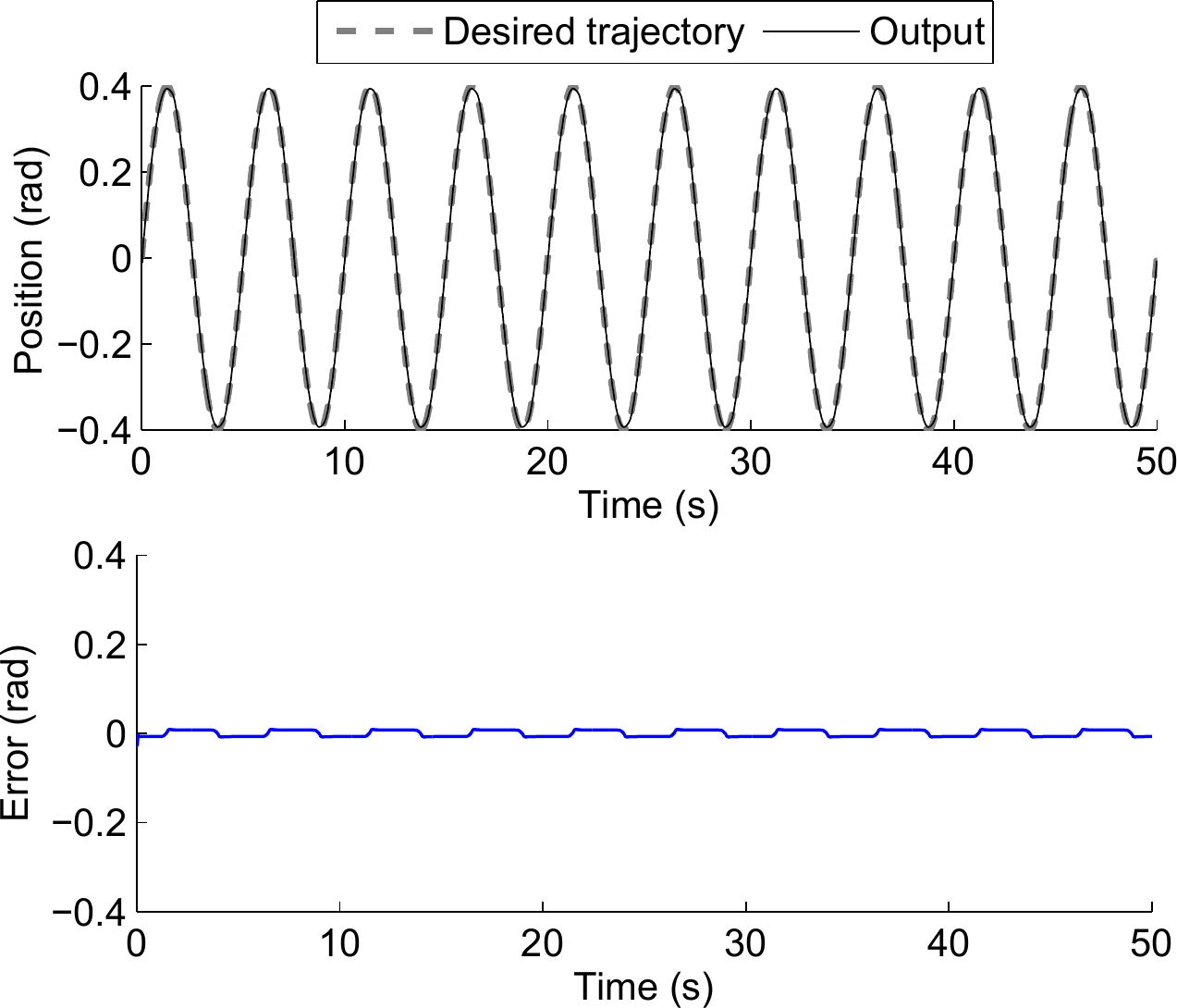}
\caption{Simulation results with the designed adaptive laws:(Upper panel) Time history of the desired trajectory $y_r$ and the output $y$; (Lower panel) Tracking error}
\label{figure3}
\end{figure}

\begin{figure}[h]
\centering
\includegraphics[width=0.36\textwidth]{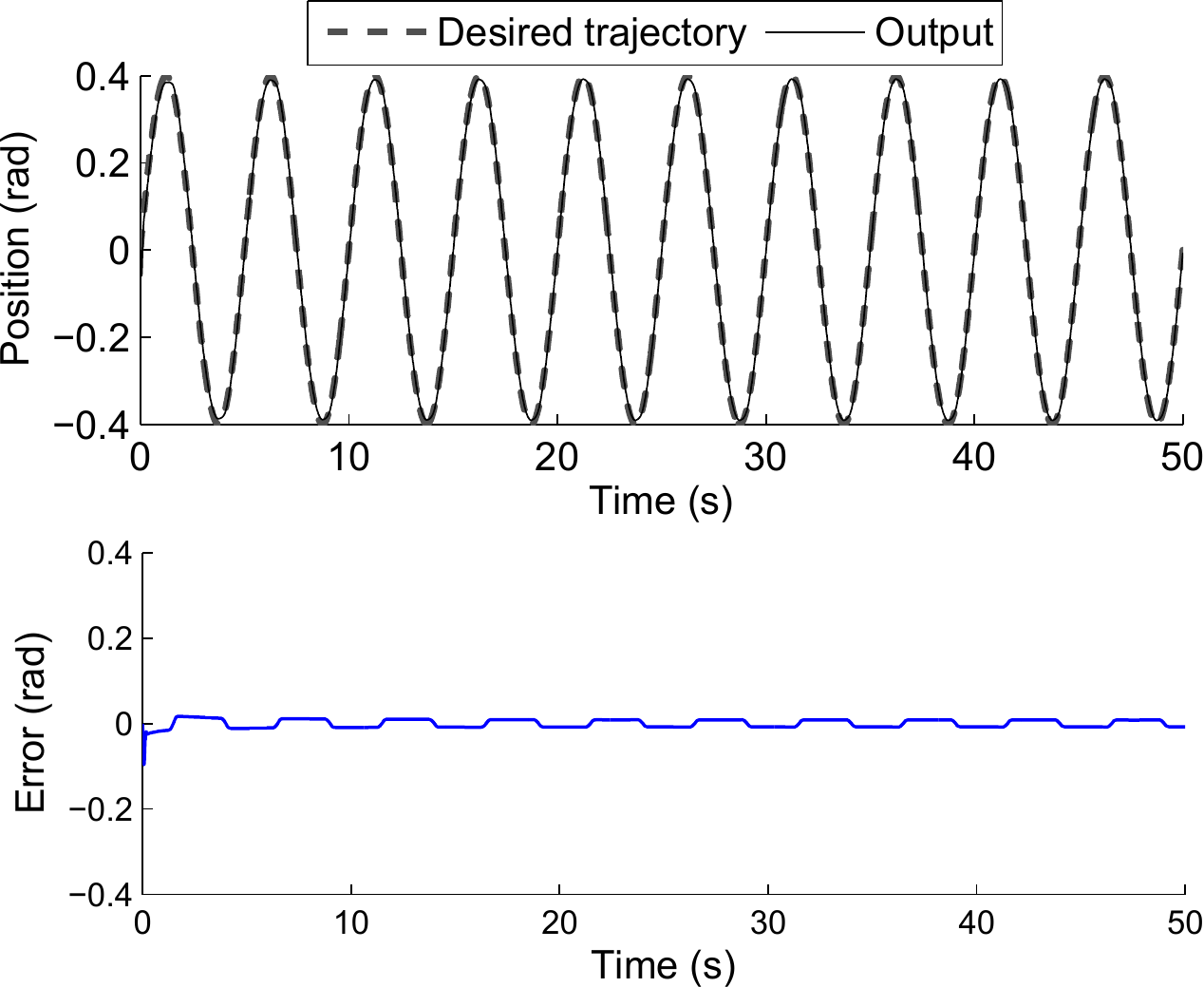}
\caption{Simulation results with the designed adaptive laws:(Upper panel) Time history of the desired trajectory $y_r$ and the output $y$; (Lower panel) Tracking error}
\label{figure4}
\end{figure}

On the basis of the chosen parameters, one of the simulation results for the designed control laws is presented. Fig. \ref{figure2} shows the control responses for the adaptive schemes given by Eq. (\ref{eq10}) to Eq. (\ref{eq14}). There is a good transient performance for the tracking error. The mean square error ($MSE$) will be also used and the tracking error posseses $MSE=4.3855\times 10^{-6}$. It is noted that the tracking error can be reduced by increasing the designed gains $\alpha_{v1},~\alpha_{v2}$. However, large values of these gains can cause vibration in the actuation device (motor). In addition, it can be also observed that the tracking error can not be eliminated completely. The main reason is that the tracking error is proven to be uniformly ultimately bounded and it is kept to stay inside a compact region $\Omega$ as $t\to \infty $ (See \textit{Theorem} 1 for more details). The parameters $\sigma_1, \sigma_2, \sigma_3$ are used to prevent the estimate values of $\hat{\Theta}, \hat{m}, \hat{D}^*$ from becoming very large. However, large values of $\sigma_1, \sigma_2, \sigma_3$ can lead to a higher value for the tracking error. Hence, small values for $\sigma_1, \sigma_2, \sigma_3$ are expected. Fig. \ref{figure3} shows the simulation results for the adaptive control laws with the same parameters as in the previous case but the values of $\sigma_1, \sigma_2, \sigma_3$ are assigned to 0.1, 0.1, and 0.1, respectively. It can be seen that there are higher values of the tracking error ($MSE=5.0782\times 10^{-5}$) compared to the one but with smaller values of $\sigma_i=0.01~(i=1,2,3)$. Fig. \ref{figure4} presents the simulation results for the case where all the model and adaptive parameters are kept the same as the first simulation but the smaller values of  
$\alpha_{v1}=3,~\alpha_{v2}=8,~k_{\Theta}=0.25,~k_m=0.25,~k_D=0.5$. We can see that a higher error of $MSE=9.4264\times 10^{-5}$ compared to the first case where $MSE=4.3855\times 10^{-6}$.

\section{Experimental Validations}
In this section, the proposed controller is further validated on a real experimental system.
\subsection{Experimental setup}
\begin{figure}
\centering
  \mbox{\subfloat[]{\label{subfig:a} \includegraphics[width=0.45\textwidth]{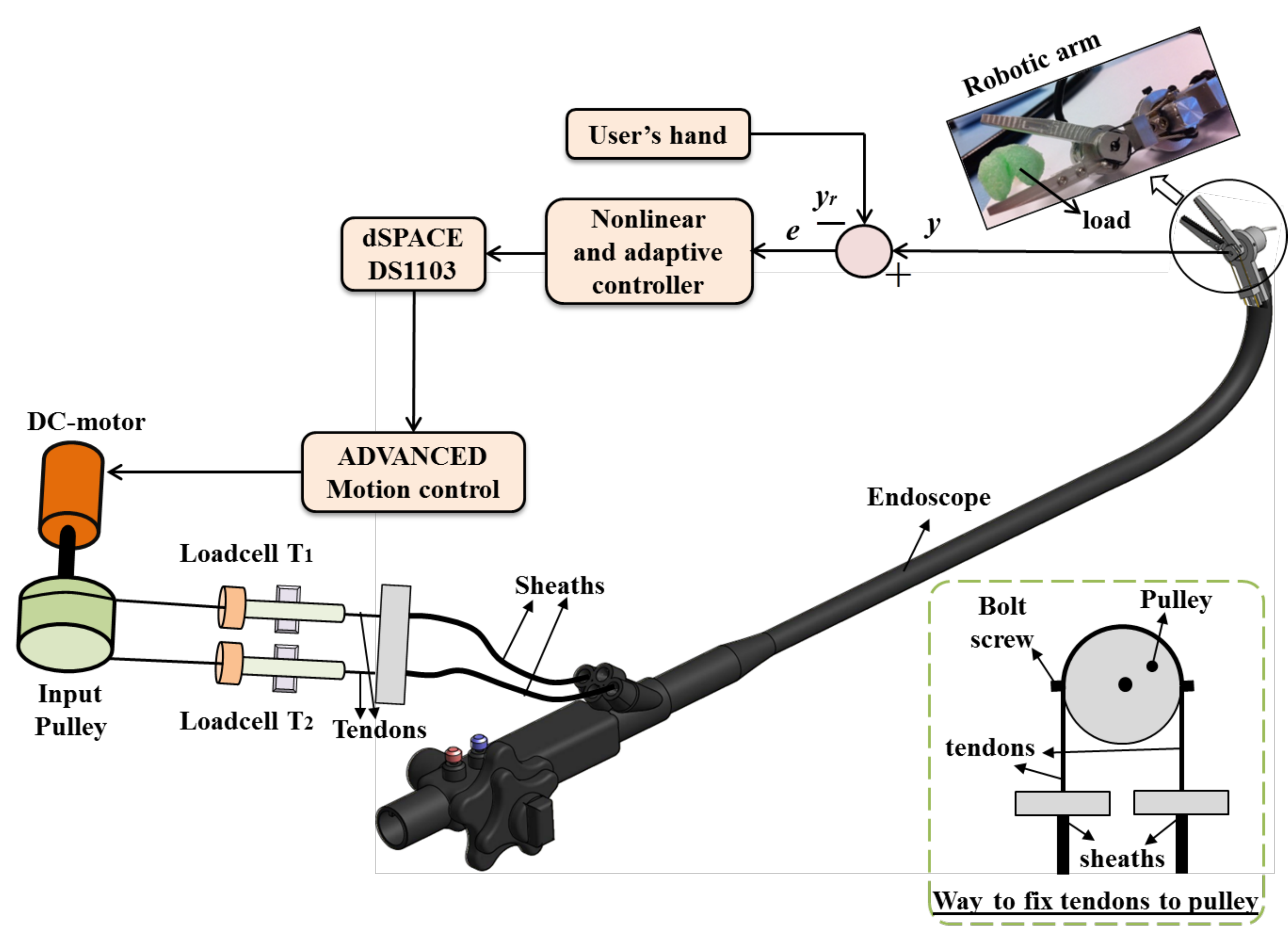}}}
    \mbox{\subfloat[]{\label{subfig:b} \includegraphics[width=0.40\textwidth]{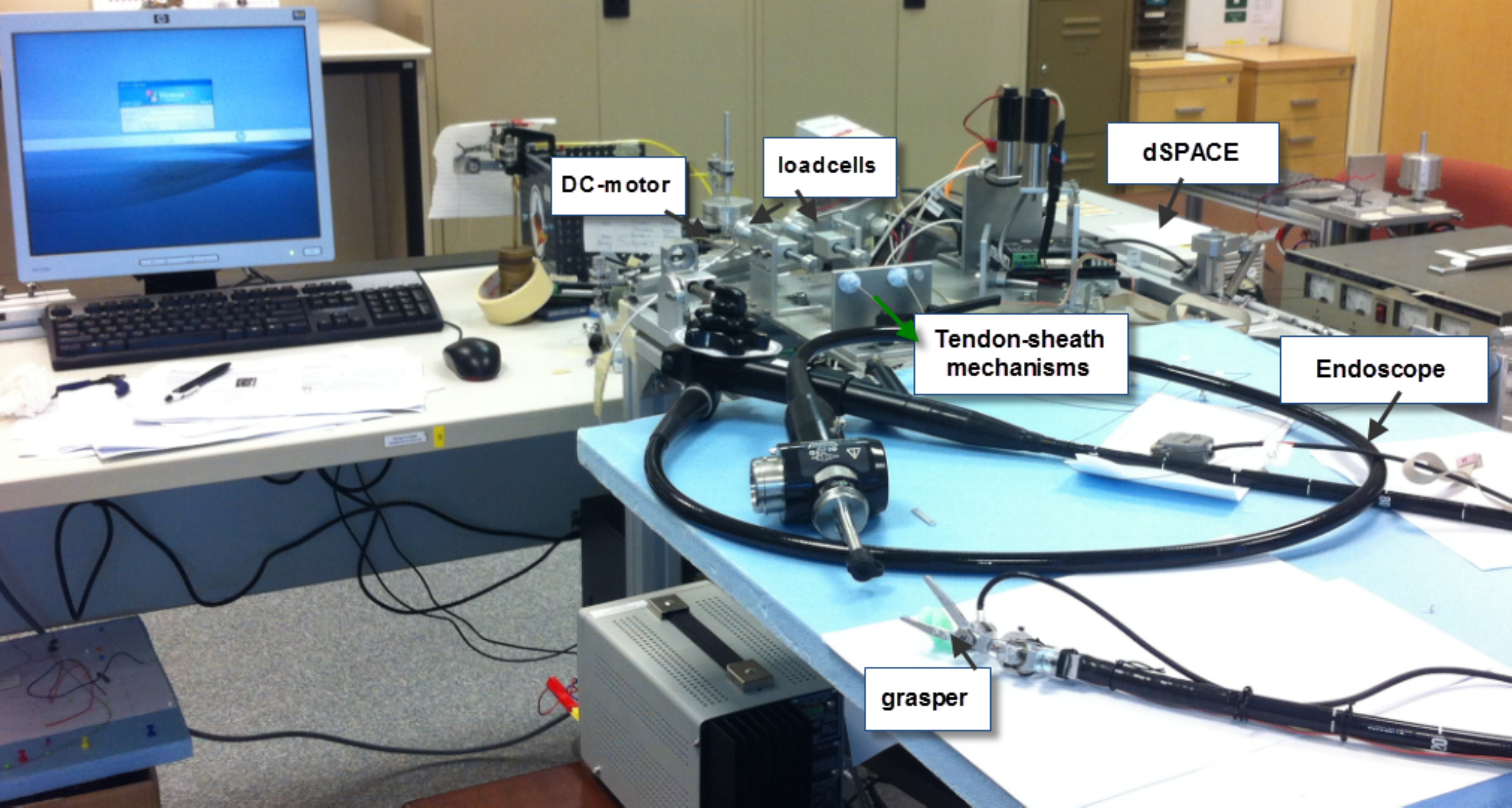}}}
    \caption{Experimental setup: (a) Illustrated diagram of the experimental platform, (b) Photo}
    \label{figure5}
\end{figure}

An experimental platform is presented in this section to demonstrate the capabilities of the proposed controller and updated laws in practical implementation. It consists of a DC-motor (Faulhaber 3863-024C) as an external actuator, a pair of TSMs, a flexible endoscope, two loadcells integrated on frictionless sliders, and a single DOF robotic arm mounted on the endoscope tip. The TSM which is provided by Asahi Intecc Co. consists of a wire cable with a coated teflon outside WR7x7D0.27mm and a round wire coil sheath with inner diameter and outer diameter of 0.36mm and 0.8mm, respectively. The length of the TSM is around 1.5m. The endoscope is a type of GIF-1T160 from Olympus, Japan. Two loadcells LW-1020-50 from Interface Corporation are used to monitor the output torque generated by the external actuator. Each tendon is connected to relevant pulleys (an input pulley at the actuator side and an output pulley at the robotic arm side-See Fig. \ref{figure5}) by bolt screws. The tendons are routed along the sheaths where each sheath is fixed on the two blocks in order to prevent their sliding during the experiment process. The output angular position $y$ from the robotic arm is measured using the high resolution encoder SCA16 from SCANCON. The input and feedback signals are decoded using DS1103 controller from dSPACE and the MATLAB Simulink from MathWorks. In the experimental validation, the DC-motor is run in current control mode and the robotic arm is kept to touch an elastic object (environmental load). The output position feedback from the robotic arm is sent to the dSPACE controller and the designed controller $u$ will be generated by the dSPACE and will be sent to the DC-motor via the ADVANCED motion control device. The two loadcells which provide the tension information $T_1, T_2$ are used to monitor the output torque generated by the motor. During the experiment, the flexible endoscope will be fixed and/or varied its configuration in order to examine the robustness of the designed controller and adaptive laws.

\subsection{Velocity estimation}
\begin{figure}[h]
\centering
\includegraphics[width=0.36\textwidth]{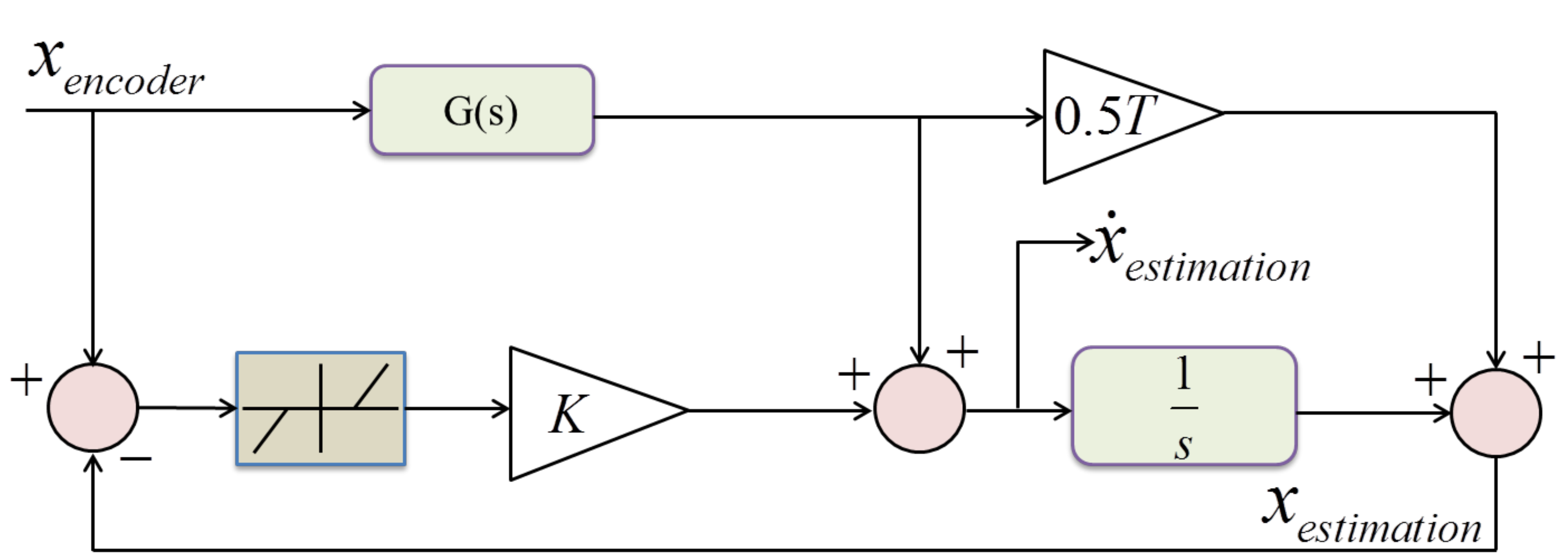}
\caption{Velocity estimator from the measured position of encoder, reproduced from \cite{noiseestimatorTilli}}
\label{figure6}
\end{figure}

The proposed adaptive controller requires the information of angular position and velocity from the output pulley. It has been known that the velocity information is often obtained by numerical differentiation of the quantized position signal from the encoder. However, the discontinuities in the encoder signal result in large spikes in the acquired velocity. The simplest method for velocity estimation from the position encoder signals is the use of standard techniques like lowpass filter or Euler approximation. However, these methods introduce significant delay. There are several methods to provide an accurate estimation of the velocity from the measured position of the encoder in the literature \cite{noiseestimatorSteinbuch}, \cite{1023168}, \cite{6428719}. Among all of them, a low-noise velocity estimation offers simple approach, high efficiency, and easy to implement for estimating the velocity to the controller design \cite{noiseestimatorTilli}. In this method, only the measured position from the encoder is required. The block diagram for velocity estimator is presented in Fig. \ref{figure6}.

\begin{figure}[h]
\centering
\includegraphics[width=0.35\textwidth]{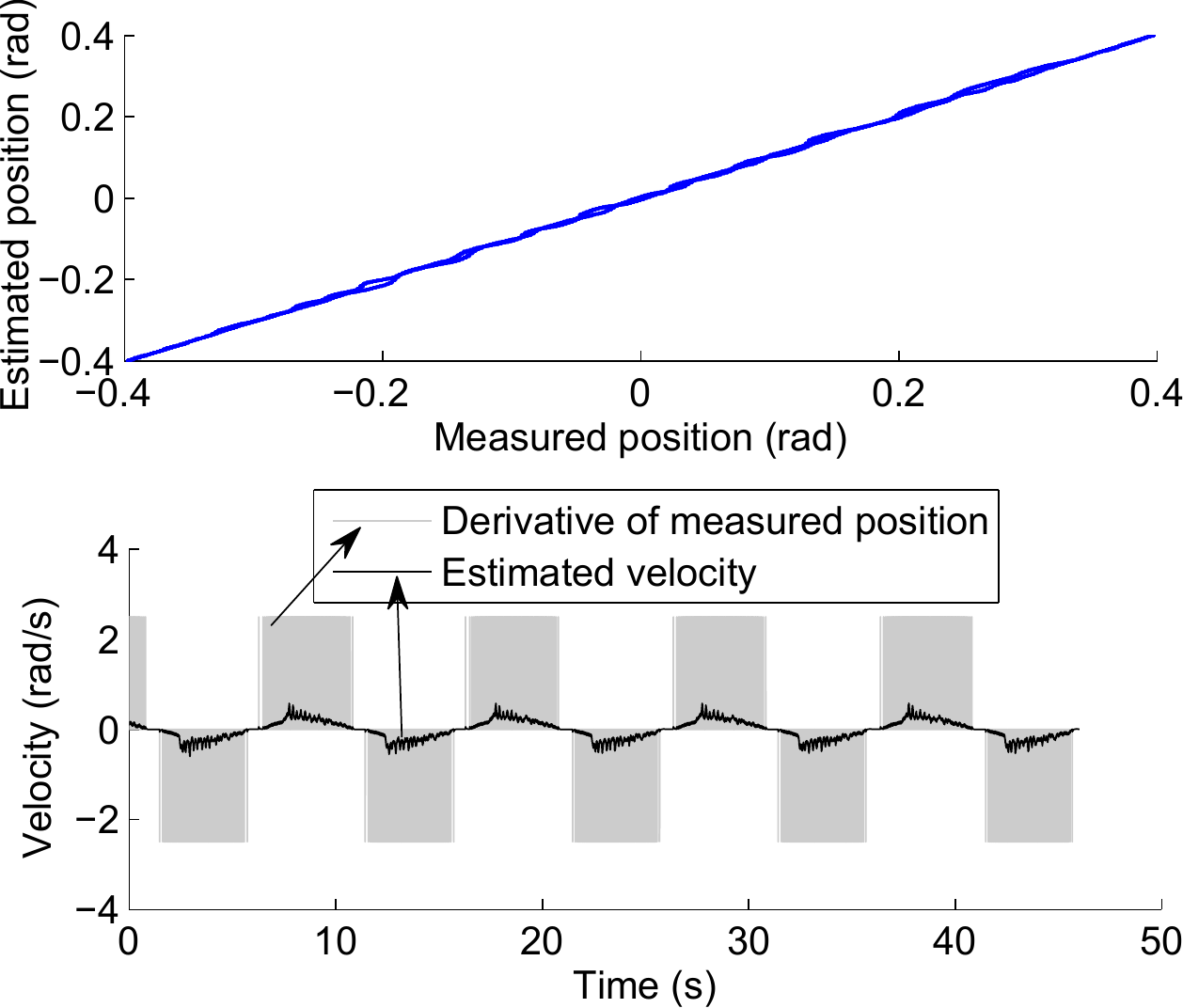}
\caption{Velocity estimator from the measured position of encoder}
\label{figure_velocity}
\end{figure}

In Fig. \ref{figure6}, $x_{encoder}$ and $x_{estimation}$ are the measured position from the encoder and the estimated position after using the proposed filter, respectively; $\dot{x}_{estimation}$ is the estimate of velocity from the position encoder; the transfer function $G(s)=s/(\tau s+1)$ where parameter $\tau$ is the filter time-constant. The gain $K$ determines the fast part of the speed estimator. The dead-zone is determined basing on the angular resolution of the encoder where the deadband $\Delta x =\pi/2n$ and $n$ is number of cycles per revolution of the encoder. $T$ is the sampling time for the experimental process. In the experiment, the parameter for velocity estimation are $T=0.01, \tau=0.2, K=20, n=3600$. The experimental result for the real differentiation of the measure signal from the encoder and by the proposed velocity estimate is shown in Fig. \ref{figure_velocity}. The upper panel of this figure presents relation between the $x_{encoder}$ and $x_{estimation}$ while the lower panel of this figure introduces the estimation result for the velocity.

\subsection{Experimental validation results}

\begin{figure}[h]
\centering
\includegraphics[width=0.36\textwidth]{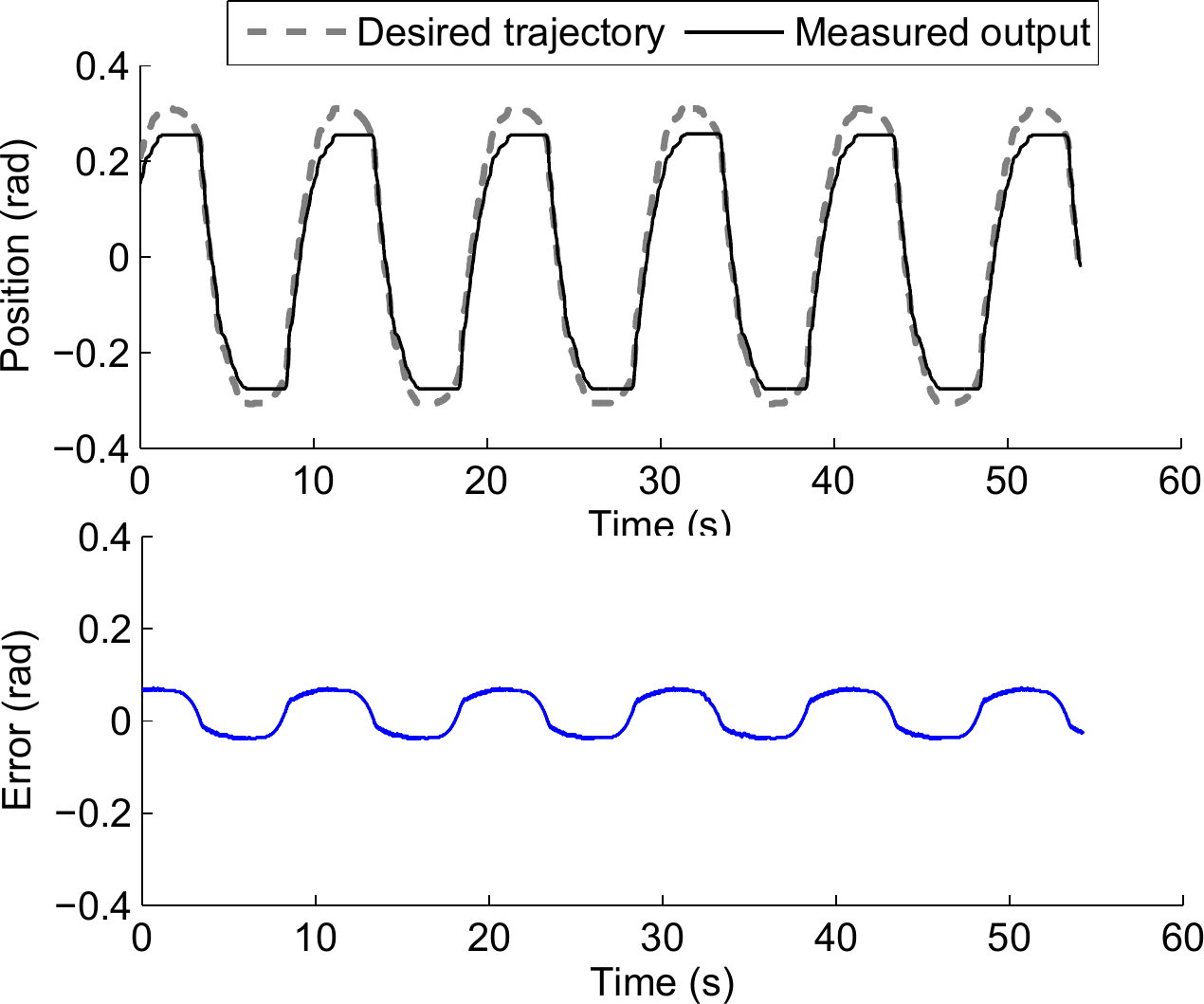}
\caption{Tracking performance of the system without using any compensation control schemes: (Upper panel) Time history of the desired trajectory $y_r$ and the output $y$; (Lower panel) Tracking error}
\label{figure7}
\end{figure}

Experiments have been carried out using the designated setup given by section V-$A$. On the basis of the designed control and updated parameters given by section IV, the nonlinear and adaptive controller (\ref{eq10}) to (\ref{eq14}) can be easily implemented in real-time experiments. In order to demonstrate the effectiveness of the proposed control scheme, five trials will be carried out for the experiments. For illustration, one of the trials will be presented only. Fig. \ref{figure7} shows the experimental results of one of the five trials without using any compensation control schemes. The upper panel of this figure introduces the time history of the desired trajectory $y_r=0.3\sin(0.4\pi t)$. and the measured position output $y$. It can be observed that the output $y$ always lags behind the desired trajectory $y_r$ with a high error value of mean square error ($MSE$) of 0.0022 rad$^2$. For easier observation and comparison, the tracking error between $y$ and $y_r$ is also shown in the lower panel of Fig. \ref{figure7}.

\begin{figure}[h]
\centering
\includegraphics[width=0.36\textwidth]{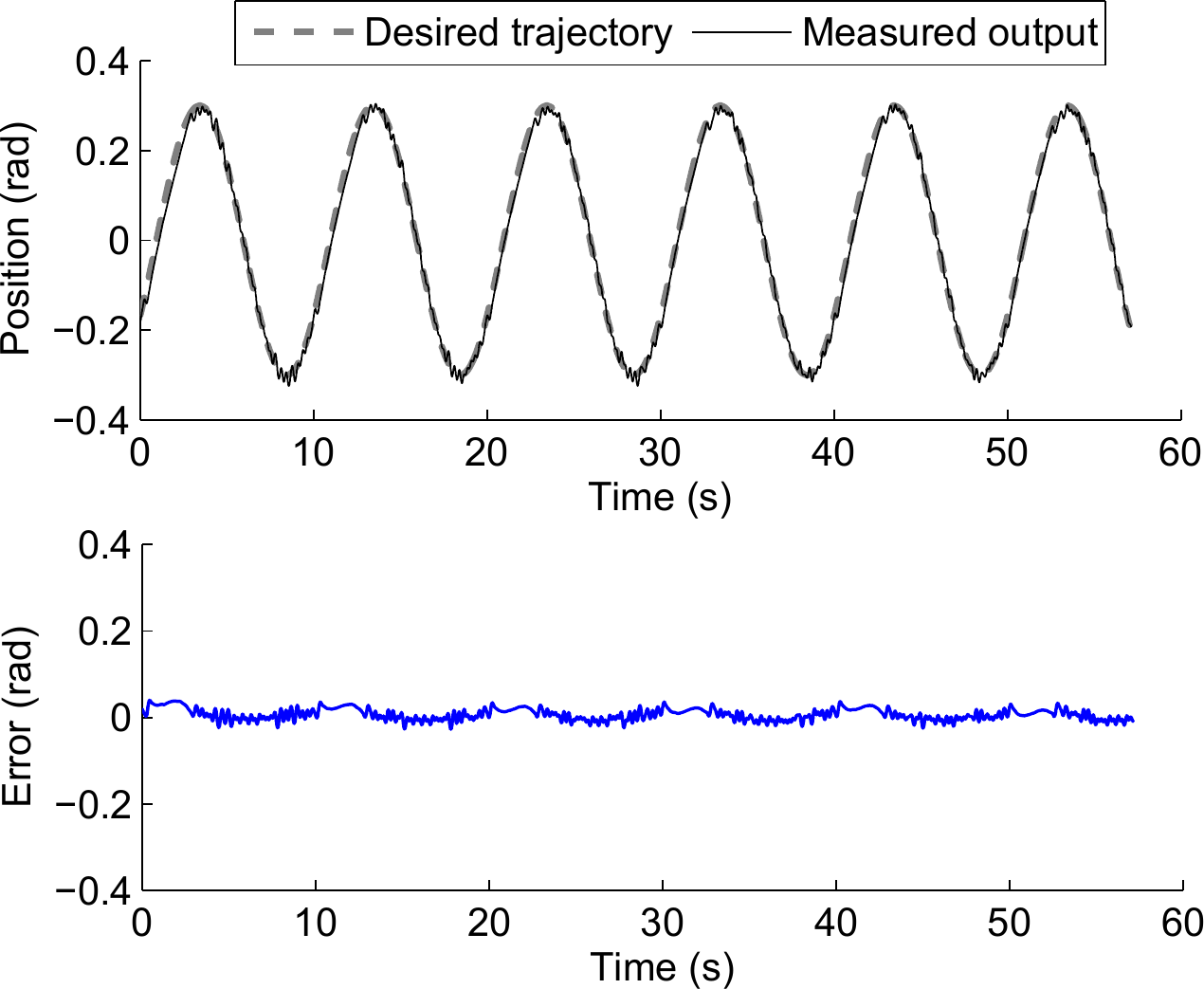}
\caption{Experimental results for the proposed adaptive controller with fixed the endoscope configuration: (Upper panel) Time history of the desired trajectory $y_r$ and the output $y$; (Lower panel) Tracking error}
\label{figure8}
\end{figure}

\textit{For the case of fixed configuration of the endoscope}: The endoscope is kept at a fixed shape configuration during the experiment. In this experiment, the nonlinear and adaptive controller will drive the measured output $y$ to accurately follow the desired trajectory $y_r=0.4\sin(0.4\pi t)$. Fig. \ref{figure8} presents the experimental result for this case. It can be observed that the measured output position $y$ accurately tracks the desired path $y_r$ (see the upper panel of Fig. \ref{figure8}). There is a significant decrease in the tracking error from a high value of $MSE=0.0022$ rad$^2$ before compensation to a smaller value of $MSE=2.1977\times 10^{-4}$ rad$^2$. This reduction can be easily seen from the lower panel of Fig. \ref{figure8}. Quantitative comparison measures of the validation are shown in Table \ref{table1}.

\begin{figure}[h]
\centering
\includegraphics[width=0.36\textwidth]{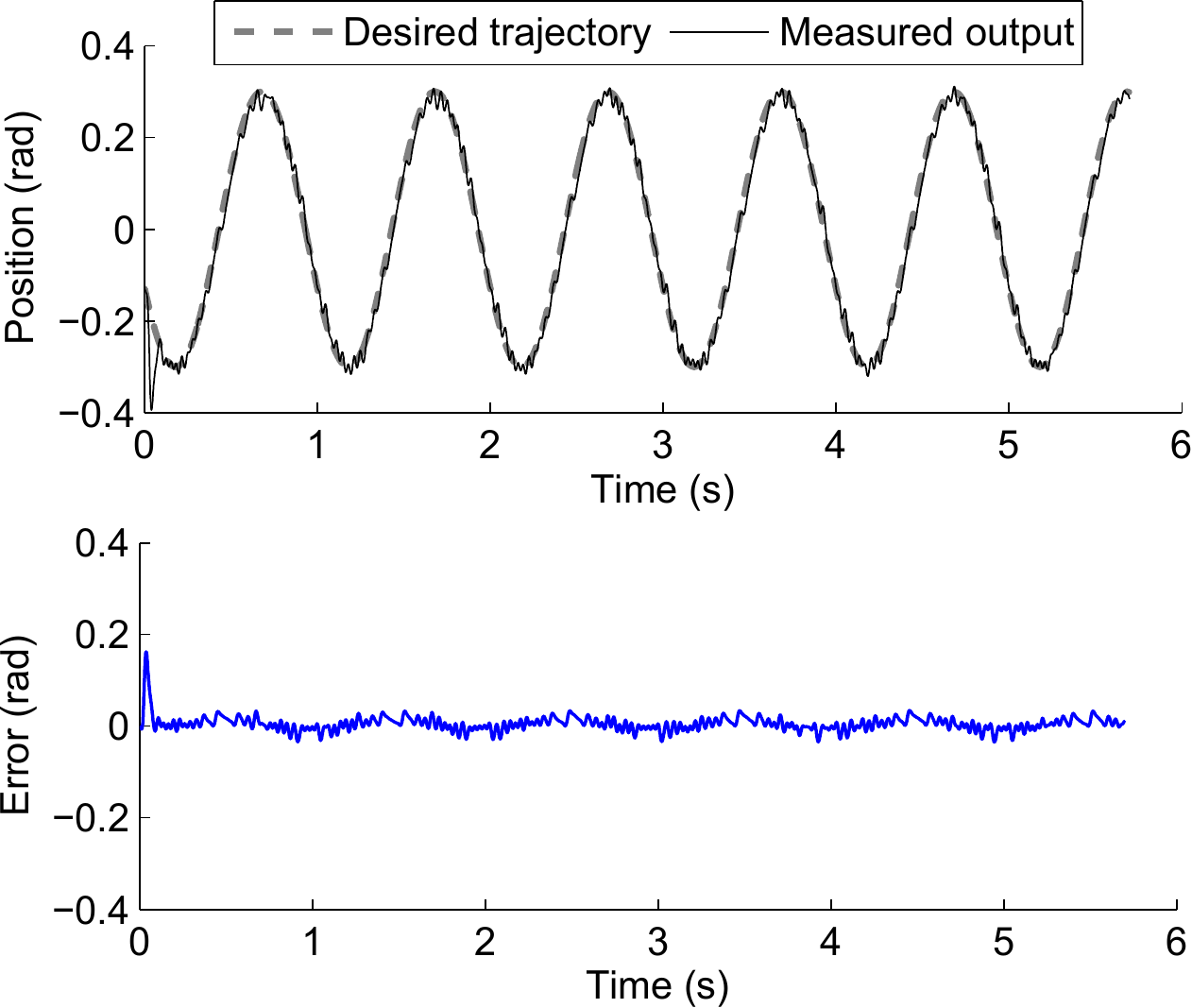}
\caption{Experimental results for the proposed adaptive controller with variation of the endoscope configuration: (Upper panel) Time history of the desired trajectory $y_r$ and the output $y$; (Lower panel) Tracking error}
\label{figure9}
\end{figure}

\begin{table}[h]
\centering
\caption{Quantitative measures of the error using the designed controller}
\label{table1}
\begin{IEEEeqnarraybox}[\IEEEeqnarraystrutmode\IEEEeqnarraystrutsizeadd{2pt}{1pt}]{c/c/c/c/c/c/c}
\hline\hline
&\mbox{Trials}&& MSE~(\text{Fixed configuration}) && MSE~(\text{Varied configuration}) &\\
\IEEEeqnarrayrulerow\\ 
\IEEEeqnarrayseprow[3pt]\\ 
& 1 && 2.1977\times 10^{-4} && 2.7876\times 10^{-4} &\IEEEeqnarraystrutsize{0pt}{0pt}\\
\IEEEeqnarrayseprow[3pt]\\
& 2 &&1.1921\times 10^{-4} && 2.1214\times 10^{-4} &\IEEEeqnarraystrutsize{0pt}{0pt}\\
\IEEEeqnarrayseprow[3pt]\\
& 3 &&2.5114\times 10^{-4} && 2.8328\times 10^{-4} &\IEEEeqnarraystrutsize{0pt}{0pt}\\
\IEEEeqnarrayseprow[3pt]\\
& 4 &&2.3282\times 10^{-4} && 2.9655\times 10^{-4} &\IEEEeqnarraystrutsize{0pt}{0pt}\\
\IEEEeqnarrayseprow[3pt]\\
& 5 &&1.6461\times 10^{-4} && 3.3967\times 10^{-4} &\IEEEeqnarraystrutsize{0pt}{0pt}\\
\IEEEeqnarrayseprow[3pt]\\
\IEEEeqnarraydblrulerow 
\end{IEEEeqnarraybox}
\end{table}
\textit{For the case of varied configuration of the endoscope}: The previously presented validation has not demonstrated the fundamnetal improvement of the proposed control scheme when the endoscope varies during the experiment. As mentioned in the first section, the friction force varies with the change of the endoscope configuration. In this section, the configuration of the endoscope is randomly varied during the validation. It can be seen that the designed controller is able to deal with the disturbance due to the change of the endoscope configuration. The upper panel of Fig. \ref{figure9} shows that there is a good tracking performance between $y_r$ and $y$ when the nonlinear controller is applied to the system. A significant decrease of the tracking error from $MSE=0.0022$ rad$^2$ to a smaller value of $MSE=2.7876\times 10^{-4}$ rad$^2$. There is a small change in the tracking error compared to the case where no change of the endoscope configuration is carried out. The quantitative measures in terms of $MSE$ are also introduced in the Table \ref{table1}. 

\section{Discussions and Conclusion}
An adaptive controller for a friction model in the TSM has been developed. Simulation and experimental tests have been carried out to show the better performances of the designed adaptive scheme. The proposed approach is able to deal with nonlinearities in the presence of unknown friction model parameters and unknown environments. Unlike current approaches on the TSM control in the literature, our proposed scheme has effectively reduced the tracking error and it is proven to be robust. Comparisons between the experimental results and the simulation show good performances. No exact knowledge of friction and backlash hysteresis parameters are required. This means that the proposed approach has potential benefits to real implementations for the TSM. Although the proposed model works well for both simulation and real-time experiment, traditional encoders have been used to get the position feedback from the distal end. It has been known that some applications can utilize advanced techniques such as 3D ultrasound probe or image-based methods to obtain the position output feedback and reduce the bulky space for the systems \cite{BardouImprovement}, \cite{KesnerIEEE},\cite{reilink2013image}. Hence, these non-conventional techniques can be applied to the system in order to fulfill the needs for the real applications. In addition, there still exists a little vibration in the experimental results. This vibration may come from some unexpected unknown disturbances and the friction in the DC-motor during the experiment. The sensor noise may also be a part of the issue. Therefore, in the future work, we will take into account the dynamic model for the motor under the presence of unknown friction and backlash hysteresis. New nonlinear and adaptive control schemes will be developed to accurately estimate and identify all the unknown models and parameters. Higher efficiency for the velocity estimation from the measured encoder will be also developed. In-vivo on live animal and human will be also carried out to higher degrees of freedom of the slave manipulator for further validations. 
\section*{Acknowledgment}
The authors would like to thank Mr. Dung Van Than for his help to calibrate devices and setup experimental process.
\appendices
\section{Proof of Theorem 1}
Define a Lyapunov function candidate $V$ as follows:
\begin{align}
V=\frac{\xi_1^2}{2}+\frac{\xi_2^2}{2}+\frac{\tilde{\Theta}^T\tilde{\Theta}}{2k_\Theta}+\frac{\tilde{m}^2}{2mk_m}+\frac{(\tilde{D}^*)^2}{2mk_D}                                                                           \label{eqA1}
\end{align}

The derivative of $V$ along (\ref{eq8}) to (\ref{eq14}) can be obtained as:
\begin{align}
&\dot{V}=\xi_1\dot{\xi}_1+\xi_2\dot{\xi}_2-\frac{\tilde{\Theta}^T\dot{\hat{\Theta}}}{k_{\Theta}}-\frac{\tilde{m}\dot{\hat{m}}}{m k_{m}}-\frac{\tilde{D}^*\dot{\hat{D}}^*}{mk_{D}}=\xi_1(\xi_2-\alpha_{v1}\xi_1) \nonumber\\
&+\xi_2(\frac{1}{m}((r_o/r_i)u+D)+\Theta^T\varphi-\ddot{y}_r+\alpha_{v1}\dot{\xi}_1)-\frac{\tilde{m}\dot{\hat{m}}}{m k_{m}}\nonumber\\
&-\frac{\tilde{\Theta}^T\dot{\hat{\Theta}}}{k_{\Theta}}-\frac{\tilde{D}^*\dot{\hat{D}}^*}{mk_{D}}=-\alpha_{v1}\xi_1^2 
-\alpha_{v2}\xi_2^2+(\tilde{\Theta}^T\varphi\xi_2-\frac{\tilde{\Theta}^T\dot{\hat{\Theta}}}{k_{\Theta}})\nonumber\\
&-(\frac{\tilde{m}\bar{u}\xi_2}{m}+\frac{\tilde{m}\dot{\hat{m}}}{m k_{m}})+(\frac{D\xi_2}{m}
-\frac{\tilde{D}^*\dot{\hat{D}}^*}{mk_{D}}-\frac{\hat{D}^*\xi_2\tanh(\xi_2/\epsilon)}{m})
\label{eqA2}
\end{align}

With (\ref{eq12}) to (\ref{eq14}), we have:

\begin{align}
&\tilde{\Theta}^T\varphi\xi_2-\frac{\tilde{\Theta}^T\dot{\hat{\Theta}}}{k_{\Theta}}=\tilde{\Theta}^T\varphi\xi_2
-\frac{\tilde{\Theta}^T(k_{\Theta}\xi_2\varphi-\sigma_1\hat{\Theta})}{k_\Theta}=\frac{\sigma_1\tilde{\Theta}^T\hat{\Theta}}{k_\Theta}
\nonumber\\
&\frac{\tilde{m}\bar{u}\xi_2}{m}+\frac{\tilde{m}\dot{\hat{m}}}{m k_{m}}=\frac{\tilde{m}\bar{u}\xi_2}{m}
-\frac{\tilde{m}(k_m\xi_2\bar{u}+\sigma_2\hat{m})}{m k_m}=-\frac{\sigma_2\tilde{m}\hat{m}}{m k_m} \nonumber \\
&\frac{\tilde{D}^*\dot{\hat{D}}^*}{mk_{D}}=\frac{(D^*-\hat{D}^*)k_D\xi_2\tanh(\xi_2/\epsilon)}{mk_{D}}-\frac{\sigma_3\tilde{D}^*\hat{D}^*}{mk_{D}}
\nonumber
\end{align}

Then (\ref{eqA2}) can be expressed by:
\begin{align}
\dot{V}&=-\alpha_{v1}\xi_1^2-\alpha_{v2}\xi_2^2+\frac{\sigma_1\tilde{\Theta}^T\hat{\Theta}}{k_\Theta}+\frac{\sigma_2\tilde{m}\hat{m}}{m k_m}
+\frac{\sigma_3\tilde{D}^*\hat{D}^*}{mk_{D}}\nonumber\\
&\quad +\frac{D\xi_2}{m}-\frac{D^*\xi_2\tanh(\xi_2/\epsilon)}{m}\label{eqA3}
\end{align}

As introduced in \cite{RobustAdaptiveFeng}, the $\tanh(\xi_2/\epsilon)$ function obeys the following property: $|\xi_2|-\xi_2\tanh(\xi_2/\epsilon)\leq 0.2785\epsilon$. Apply the Young's inequality for two numbers $a$ and $b$,~i.e. $ab \leq 0.5(a^2+b^2)$, the third, fourth, and fifth terms in right hand side of (\ref{eqA3}) can be reformulated as:
\begin{align}
\frac{\sigma_1\tilde{\Theta}^T\hat{\Theta}}{k_\Theta}&=\frac{\sigma_1\tilde{\Theta}^T(\Theta-\tilde{\Theta})}{k_\Theta} 
\leq \frac{\sigma_1\Theta^T\Theta}{2k_\Theta}-\frac{\sigma_1\tilde{\Theta}^T\tilde{\Theta}}{2k_\Theta} \label{eqA4}\\
\frac{\sigma_2\tilde{m}\hat{m}}{mk_m}&=\frac{\sigma_2\tilde{m}(m-\tilde{m})}{mk_m} 
\leq \frac{\sigma_2 m^2}{2mk_m}-\frac{\sigma_2\tilde{m}^2}{2mk_m} \label{eqA5}\\
\frac{\sigma_3\tilde{D}\hat{D}^*}{mk_D}&=\frac{\sigma_3\tilde{D}^*(D^*-\tilde{D}^*)}{mk_D} 
\leq \frac{\sigma_3 {D^*}^2}{2mk_D}-\frac{\sigma_3(\tilde{D}^*)^2}{2mk_D} \label{eqA6}
\end{align}

With (\ref{eqA4}) to (\ref{eqA6}), $\dot{V}$ in (\ref{eqA3}) can be rewriten by:

\begin{align}
 &\dot{V}\leq -\alpha_{v1}\xi_1^2-\alpha_{v2}\xi_2^2-\frac{\sigma_1\tilde{\Theta}^T\tilde{\Theta}}{2k_\Theta}-\frac{\sigma_2\tilde{m}^2}{2mk_m}-\frac{\sigma_3(\tilde{D}^*)^2}{2mk_D}
\nonumber\\
&+\frac{\sigma_1\Theta^T\Theta}{2k_\Theta}+\frac{\sigma_2 m^2}{2mk_m}+\frac{\sigma_3 (D^*)^2}{2mk_D}+\frac{D^*}{m}(|\xi_2|-\xi_2\tanh(\frac{\xi_2}{\epsilon})) \nonumber\\
&\leq 
-\alpha_{v1}\xi_1^2-\alpha_{v2}\xi_2^2-\frac{\sigma_1\tilde{\Theta}^T\tilde{\Theta}}{2k_\Theta}-\frac{\sigma_2\tilde{m}^2}{2mk_m}-\frac{\sigma_3(\tilde{D}^*)^2}{2mk_D}
\nonumber\\
&+\frac{\sigma_1\Theta^T\Theta}{2k_\Theta}+\frac{\sigma_2 m}{2k_m}+\frac{\sigma_3 (D^*)^2}{2mk_D}+\frac{0.2785\epsilon D^*}{m}\nonumber\\
&=-\varrho V+\Psi \label{eqA7}
\end{align}
where $\Psi=(\sigma_1/2k_{\Theta})\Theta^T\Theta+0.2785\epsilon D^*/m+(\sigma_2/2k_m)m+(\sigma_3/ 2mk_D) (D^*)^2$; $\varrho=\text{min}\{2\alpha_{v1}, 2\alpha_{v2}, \sigma_1, \sigma_2, \sigma_3\}$. 

Solving $V$ from $\dot{V} \leq -\varrho V+\Psi$ and from (\ref{eqA1}), one can obtain:
\begin{align}
0\leq 0.5\xi_1^2 \leq V \leq (V(0)-\Psi/\varrho)e^{-\varrho t}+\Psi/\varrho \label{eqA8}
\end{align}
where $V(0)=0.5(\xi_1(0))^2+(\tilde{\Theta}(0))^T\tilde{\Theta}(0)/(2k_\Theta)+0.5(\xi_2(0))^2+(\tilde{m}(0))^2)/(2mk_m)
+(\tilde{D}^*(0))^2)/(2mk_D)$.

It can be seen that there exists a time $T>0$ such that $\forall t>T$, $(V(0)-\Psi/\varrho)e^{-\varrho t} \to 0$. Then $V$ is bounded by $\Psi/\varrho$ for $\forall t>T$. Hence $V$ is uniformly ultimately bounded (UUB); thus the tracking error $e_r=\xi_1, \tilde{D}^*, \tilde{m}, \tilde{\Theta}$ are also bounded. This further guarantees the boundedness of $\hat{D}^*, \hat{m}, \hat{\Theta}$ since $D^*,m,\Theta$ are bounded.

From (\ref{eqA8}), the variable $\xi_1$ can be expressed by:
\begin{align}
|e_r|=|\xi_1| \leq \sqrt{2(V(0)-\Psi/\varrho)e^{-\varrho t}+2\Psi/\varrho} \label{eqA9}
\end{align}

For $\forall t>T$, the tracking error $e_r=\xi_1$ will converge to a compact set $\Omega=\{|e_r|\big{|}|e_r| \leq 2\sqrt{\varrho/\Psi}\}$ since $(V(0)-\Psi/\varrho)e^{-\varrho t} \to 0$. Consequently, the control input $u$ is also bounded. The proof is completed here.

\ifCLASSOPTIONcaptionsoff
  \newpage
\fi


\bibliographystyle{IEEEtran}
\bibliography{Citation_Journal7}

\end{document}